\documentclass[conference]{IEEEtran}
\IEEEoverridecommandlockouts
\usepackage{cite}
\usepackage{amsmath,amssymb,amsfonts}
\usepackage{algorithmic}
\usepackage{graphicx}
\usepackage{textcomp}
\usepackage{subfigure}
\usepackage{xcolor, algorithm, booktabs}
\usepackage{bm}
\usepackage{makecell}
\usepackage{wasysym}
\newcommand*{\affmark}[1][*]{\textsuperscript{#1}}

\newcommand{\com}{\text{com}}
\newcommand{\at}{\text{at}}

\newcommand{\safe}{\text{safe}}
\newcommand{\HD}{\text{HD}}
\newcommand{\old}{\text{old}}

\usepackage{multirow} 
\allowdisplaybreaks
\def\BibTeX{{\rm B\kern-.05em{\sc i\kern-.025em b}\kern-.08em
    T\kern-.1667em\lower.7ex\hbox{E}\kern-.125emX}}
\begin{document}

\title{Imitation Learning based Alternative Multi-Agent Proximal Policy Optimization for Well-Formed Swarm-Oriented Pursuit Avoidance  \\
\thanks{This work was supported in part by the National Natural Science Foundation of China under Grants 62071425, in part by the Zhejiang Key Research and Development Plan under Grant 2022C01093, and in part by the Zhejiang Provincial Natural Science Foundation of China under Grant LR23F010005.}
}
\DeclareRobustCommand*{\IEEEauthorrefmark}[1]{\raisebox{0pt}[0pt][0pt]{\textsuperscript{\footnotesize #1}}}
\author{\IEEEauthorblockN{Sizhao Li\affmark[1], 
Yuming Xiang\affmark[1], 
Rongpeng Li\affmark[1]$^{*}$, 
Zhifeng Zhao\affmark[2], 
Honggang Zhang\affmark[2]}
\IEEEauthorblockA{
\textit{\affmark[1]College of Information Science and Electronic Engineering, Zhejiang University}
\textit{\affmark[2]Zhejiang Lab} \\
Hangzhou, 310001, Zhejiang, China \\
\{liszh5, xiangym,  lirongpeng\}@zju.edu.cn,
\{zhaozf, honggangzhang\}@zhejianglab.com}
}


\maketitle

\begin{abstract}
Multi-Robot System (MRS) has garnered widespread research interest and fostered tremendous interesting applications, especially in cooperative control fields.
Yet little light has been shed on the compound ability of formation, monitoring and defence in decentralized large-scale MRS for pursuit avoidance, which puts stringent requirements on the capability of coordination and adaptability.
In this paper, we put forward a decentralized Imitation learning based Alternative Multi-Agent Proximal Policy Optimization (IA-MAPPO) algorithm to provide a flexible and communication-economic solution to execute the pursuit avoidance task in well-formed swarm.
In particular, a policy-distillation based MAPPO executor is firstly devised to capably accomplish and swiftly switch between multiple formations in a centralized manner.
Furthermore, we utilize imitation learning to decentralize the formation controller, so as to reduce the communication overheads and enhance the scalability. 
Afterwards, alternative training is leveraged to compensate the performance loss incurred by decentralization.
The simulation results validate the effectiveness of IA-MAPPO and extensive ablation experiments further show the performance comparable to a centralized solution with significant decrease in communication overheads.
\end{abstract}


\section{Introduction}

Nowadays, cooperative control in Multi-Robot Systems (MRS) has been attracting growing interest, since robotic swarm has demonstrated great potentials in both civilian and military tasks~\cite{s22155569}.
As for decentralized large-scale MRS, it is indispensable to develop the compound ability of Formation, Monitoring and pursuit Avoidance (FMA), which is quite prevalent in real-world.
In that regard, \cite{9157980} presents a two-level flocking control system for static obstacle avoidance and targeted positions navigation. \cite{article} proposes a hybrid system for bypassing a pre-defined predator (attacker) trajectory, by integrating flocking control and Reinforcement Learning (RL), which implements invariant formation, thus lacking the essential flexibility.
Furthermore, the considered scenarios are over-simplified and far from the reality~\cite{8340193}. 
Therefore, the process of FMA should be further calibrated to be more consistent with the practice, and effective formation control belongs to an inevitable ingredient for pursuit avoidance.

Benefiting from the robust adaptability in complex environments, Multi-Agent Reinforcement Learning (MARL) algorithms have been widely used to address formation control issues in MRS systems. For example, \cite{yan2022relative} combines Multi-Agent Proximal Policy Optimization (MAPPO) \cite{yu2022surprising} with policy distillation \cite{green2019distillation} to achieve multi-formation assignment with global observations. \cite{xiang2023decentralized} enhances the formation stability in communication-limited scenarios by incorporating an attention mechanism.
Nevertheless, these formation control schemes are responsive to abrupt factors (e.g., battery life depletion) and can not be easily extended to scenarios with malicious actions of predators, which typically are un-predictable, since individual agents in the large-scale MRS can not spontaneously reach a policy consensus in a distributed manner. 
As a remedy, central scheme for formation control satisfies the demand on stability and synchronism. However, its heavy dependency on costly communications between agents and the central controller often hampers the application in large-scale MRS. 


In order to establish a consensus for formation control in a distributed manner, a cooperative sequential decision-making framework is often leveraged, and individual agents attempt to discover beneficial behavior sequences distributively. However, such an attempt is often in vain, due to the difficulty to locate useful actions from the huge-dimensional joint action space from all agents. In other words,  
current MARL algorithms encounter severe challenges owing to the extreme sparsity of meaningful rewards from the environment \cite{10.5555/3306127.3331808}.
Fortunately, Imitation Learning (IL) \cite{zhang2018pretraining}, which emulates a centralized policy-driven rational trajectories for decentralized execution, manifests itself in converging the policy. 
However, compared to a centralized policy, such an approach often degrades the performance due to the cumulative errors between practical execution trajectories and the imitated training samples \cite{20122515132876}.

In this paper, in order to enhance the flexibility of formation control in previous FMA work \cite{9157980,article} and ameliorate the practicality issue towards distributed formation control \cite{yan2022relative}, we put forward a communication-efficient algorithm named Imitation learning based Alternative Multi-Agent Proximal Policy Optimization (IA-MAPPO). 
Compared with the existing work, the contribution of our paper can be summarized as follows.
\begin{figure}[tbp]
\centering
\includegraphics[width=0.44\textwidth]{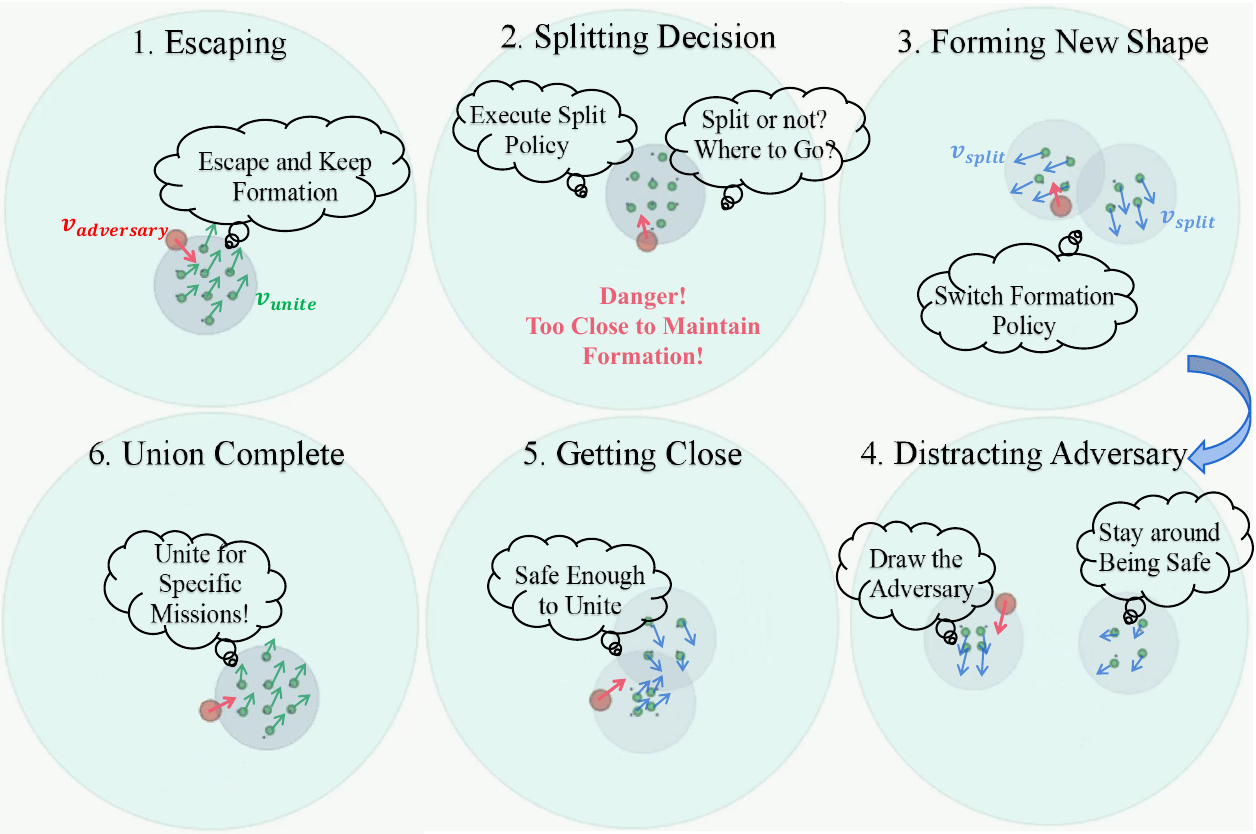}
\vspace{-0.50em}
\caption{Overview of well-formed
swarm’s pursuit avoidance with transformation from $\mathcal{F}_8$ to $\mathcal{F}_4$.}
\label{environment2}
\vspace{-1.0em}
\end{figure}

\begin{itemize}
    \item We develop a hierarchical proactive formation control framework, where the high-level policy determines the division control policy (i.e., the appropriate time for formation switching), while the low-level policy resolves to consensus-oriented distributed mixed-formation policy following our previous works \cite{xiang2023decentralized}.
    \item As for the high-level policy, we first leverage IL to learn a distributed division policy that generates the formation confidence. After communicating with neighboring agents and aggregating the confidence, observation masking is applied on individual agents to match appropriate neighbors for the next formation. Furthermore, in order to compensate the IL-induced performance degradation, we adopt Alternative Training (AT) to fine-tune low-level policy on the basis of well-trained high-level policy.
    \item We validate the effectiveness of IA-MAPPO and extensive ablation experiments further show that IA-MAPPO yields competitive performance as a centralized solution and significantly lower communication overheads.
\end{itemize}

The remainder of this paper is organized as follows. We introduce the system model and formulate the problem in Section \ref{s2}. Afterwards, we elaborate on the details of the proposed IA-MAPPO algorithm in Section \ref{s3}. In Section \ref{s4}, we present the simulation settings and discuss the experimental results. Finally, Section \ref{s5} concludes this paper.

\section{System Model and Problem Formulation}\label{s2}

\subsection{System Model}
We primarily consider a decentralized formation control problem, wherein agents in $\mathcal{N}$, with $|\mathcal{N}|=N$, are required to stay in the target area $\mathcal{T}$ in a communication-limited decentralized manner and transform within a set of pre-defined formation patterns $\{\mathcal{F}_c | \forall c \in \mathcal{C }\}$ where $c$ represents the agent quantity in formation $\mathcal{F}_c$ and $\mathcal{C}$ denotes the set of possible quantities with $|\mathcal{C}|=C$. 
At time step $t$, each agent $i$ needs to spontaneously determine one formation pattern $c$ (i.e., $\mathcal{F}_i(t)=\mathcal{F}_c$) and recognize its $c-1$ neighbors cooperatively, resulting in $\chi(t)$ groups with the number $n_i, i\in \{1,\cdots, \chi(t)\}$ of agents in each group  satisfying ${n}_{1}+\cdots+{n}_{\chi(t)}=N$. 
An overview of the process is demonstrated in Fig. \ref{environment2}, wherein $c\in \mathcal{C}=\{8,4\}$, and a formation transforms from $F_8$ to $F_4$ (implying $\chi(t)$ becomes $2$ from $1$) for pursuit avoidance.

In order to accomplish the basic FMA task for fixed formation, we formulate the problem as a Decentralized Partially Observable Markov Decision Process (Dec-POMDP) \cite{xiao2023stochastic}, which is defined as $\langle\mathcal{I}, \mathcal{S}, \mathcal{A}, \mathcal{P}, \Omega, R, \mathcal{O}, \gamma\rangle$.  In the FMA task, $\mathcal{I}$ represents $N$ agents and an adversary that utilizes default policy for pursuit. Each agent $i\in \mathcal{N}$ maintains a joint policy $\bm{\pi}=\{\pi_f,\bm{\pi}_i\}$ where $\pi_f$ denotes the division policy and $\bm{\pi}_i=\{\pi_i^c | \forall c \in \mathcal{C}\}$ represents the policy set for predefined formations.
$\mathcal{S}$ denotes the global state space while $\mathcal{A}$ is the homogeneous action space for a single agent. 
Owing to the scant
ability of perception against the colossal environment, each agent $i$ obtains a local observation $\textbf{o}_{i} \in \Omega$ via the observation function $\mathcal{O}\left(\mathbf{o}_{i} \mid \mathbf{s}, i\right): \mathcal{S} \times \mathcal{N} \times \Omega \rightarrow[0,1]$ instead of the state $\mathbf{s}\in \mathcal{S}$ at each time-step and adopts an action $\textbf{a}_{i} \in \mathcal{A}$ according to mixed-formation policy $\bm{\pi}_{i}\left(\cdot \mid \textbf{o}_{i}\right): \Omega \times \mathcal{A} \rightarrow[0,1]$.
The joint action $\textbf{a}=\left\{\textbf{a}_{1}, \textbf{a}_{2}, \cdots, \textbf{a}_{N}\right\}$ taken at the current state $\textbf{s}$ makes the environment transit into the next state $\textbf{s}'$ according to the function $\mathcal{P}\left(\textbf{s}^{\prime} \mid \textbf{s}, \textbf{a}\right): \mathcal{S} \times \mathcal{A} \times \mathcal{S} \rightarrow[0,1]$. All agents share a global reward function $R(\textbf{s}, \textbf{a}): \mathcal{S} \times \mathcal{A} \rightarrow \mathbb{R}$ with a discount factor $\gamma$. Consistent with the Dec-POMDP framework, we specify the elements as follows.

\subsubsection{State and Observation}
The raw information of agent $i$ is denoted as $\textbf{o}_i^{-}(t)=[ \textbf{p}_{i}{(t)}, \textbf{v}_{i}(t), \textbf{p}_e(t)]$ which encompasses position $\textbf{p}_i{(t)}=[p_{x_i}{(t)},p_{y_i}{(t)}]$, velocity $\textbf{v}_i{(t)}=[v_{x_i}{(t)},v_{y_i}{(t)}]$ of agent $i$ and the position of adversary $\textbf{p}_e(t)=[p_{x_e}{(t)},p_{y_e}{(t)}]$ gained from sensors.
After communicating with neighbors, agent $i$ augments its observation by $\textbf{o}_{\bm{\xi}_i}^{-}(t)$, where $\textbf{o}_{\bm{\xi}_i}^{-}(t)=\{(\textbf{p}_j{(t)},\textbf{v}_j{(t)})| \forall j\in \bm{\xi}_i{(t)}\}$ contains position and velocity of neighbors in $\bm{\xi} _i{(t)}$, who are within agent $i$'s communication range $\delta_{\com}$ (i.e., $\|\textbf{p}_i{(t)}-\textbf{p}_j{(t)}\| < \delta_{\com}$), where $\|\cdot \|$ denotes an Euclidean norm operator. 
Thus, the observation of agent $i$ is summarized as $\textbf{o}_i(t)=[\textbf{o}_i^{-}(t),\textbf{o}_{\bm{\xi}_i}^{-}(t)]$.
On the other hand, the global state includes the positions and velocities of all agents and the adversary, that is, $\textbf{s}{(t)}=[\{ \textbf{o}_i^{-}(t)| \forall i\in \mathcal{N} \},\{\textbf{p}_e{(t)},\textbf{v}_e{(t)}\}]$.

\subsubsection{Action}
Based on the local observation $\textbf{o}_i(t)$, each agent sets its acceleration $\textbf{a}_i{(t)}=[{a}_{x_i}{(t)},{a}_{y_i}{(t)}] \in \mathcal{A}$ following policy $\pi_{i}^c\left(\cdot \mid \textbf{o}_{i}\right)$ individually to complete FMA task in a $\pi_f(t)$-determined formation $\mathcal{F}_{i}(t)=\mathcal{F}_c$\footnote{We slightly abuse the terminology here, as the action shall implicitly involves the division control policy $\pi_f$ as well.}. Later in Section III, we shall further discuss the determination of $\bm{\pi}$ within the framework of IA-MAPPO.

\subsubsection{Reward}
In this part, we introduce the reward designed for a specific number $c$ of swarm in FMA task.
\begin{figure*}[tbp]
    \vspace{-2.5em}
    \centering 
    \includegraphics[scale =0.4]{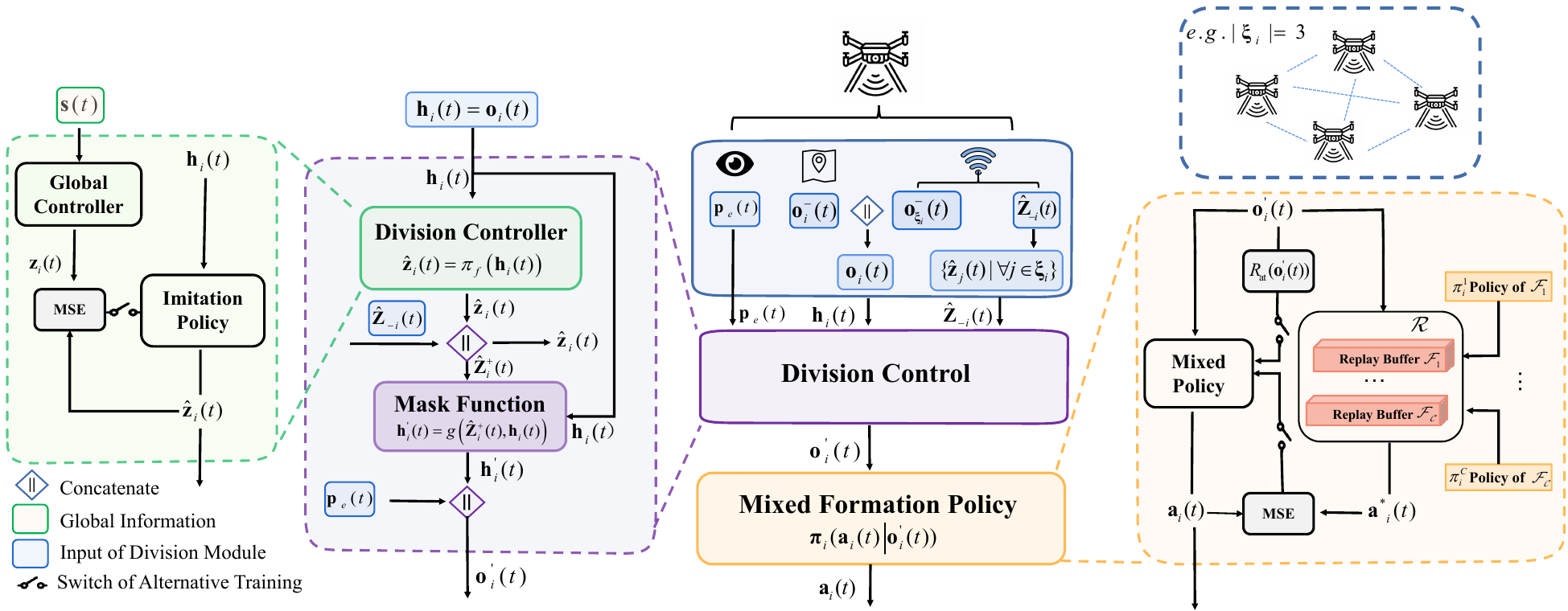}
    \vspace{-0.8em}
    \caption{The illustration of {IA-MAPPO} with Policy Distillation, Imitation Learning and Alternative Training.}  
    \label{AI-MAPPO}
    \vspace{-1.8em}

\end{figure*} 

\begin{itemize}
    \item \textbf{Formation reward}.
    Towards implementing leader-free formation control and enhancing the robustness, we adopt the Hausdorff Distance (HD) \cite{pan2022flexible} to measure the topology distance between the current and the expected formation. We denote the relative positions of agents as $\textbf{P}{(t)}=\{\textbf{p}_j{(t)}-\bar{\textbf{p}}{(t)}|\forall j \in \mathcal{F}_c\}$ where $\bar{\textbf{p}}{(t)}$ is the center of the swarm and $\mathcal{F}_c$ is the target formation corresponding to $c$. Thus the formation reward can be given by
    \begin{equation}
        \label{eq:rform}
        R_f(t) = -k_f \max \{(d_{\HD}(\textbf{P}{(t)},\mathcal{F}_c) - \varphi_f, 0\},
    \end{equation}
    where $k_f$ denotes the coefficient of formation punishment and the HD between current and ideal topology is defined as $d_{\HD}(\textbf{P}{(t)},\mathcal{F}_c) = \max \left \{h_d(\textbf{P}{(t)},\mathcal{F}_c),h_d(\mathcal{F}_c,\textbf{P}{(t)}) \right \}$ with $h_d(\textbf{P}{(t)},\mathcal{F}_c) = \mathop{\max}\nolimits_{\mathbf{x}\in \textbf{P}(t)}\mathop{\min}\nolimits_{\textbf{y}\in \mathcal{F}_c} \|\textbf{x}-\textbf{y}\|$. The hyper-parameter $\varphi_f $ is used to fine-tune the trade-off between the formation accuracy and convergence rapidity.
    \item \textbf{Target area reward}.
    In order to maintain the swarm in target area, we assume the target area $\mathcal{T}$ as a circular region with radius $r_{\mathcal{T}}$, the center of which is set at the original point (i.e., $\mathcal{T}:\{(x, y) \mid \|(x, y) \| \leqslant r_{\mathcal{T}}\}$), and the target area reward can be formulated as
    \begin{equation}
        R_{a}(t)  = \sum\nolimits_{i \in \mathcal{N}} R_{a_i}(t),
    \end{equation}
    where $R_{a_i}(t)$, which equals $-k_{a} e^{\alpha\| \textbf{p}_{i}(t)\|}$ if $\left\|\textbf{p}_{i}(t)\right\| >	 r_{\mathcal{T}}$ and nulls otherwise, represents the punishment for the agent $i$ exceeding the circle. $k_a$ and $\alpha$ are both constants.
    \item \textbf{Evasion reward}.
    In order to keep agents away from the adversary, a penalty would be added as the evasion reward if any agent collides with the adversary,
    \begin{equation}
        \label{reward sum}
        R_{e}{(t)}=-k_e\sum\nolimits_{i \in \mathcal{N}} (\| \textbf{p}_i{(t)}-\textbf{p}_e{(t)} \|<\delta _{\safe}),
    \end{equation}  
    where $k_e$ is the coefficient of evasion reward and $\delta _{\safe}$ is the minimum safe distance between agents and the adversary.
\end{itemize}

To conclude, the final reward function is as
\begin{equation}
    \label{total rewards}
    R(t) =  R_{f}{(t)}+ R_{a}{(t)} + R_{e}{(t)}.
\end{equation}

\subsection{Problem Formulation}\label{problem formulation}
Without loss of generality, at time-step $t$, each agent should unanimously determine the swarm division and decide the belonging group. 
Dependent on the centralized or decentralized controller, the algorithm takes either the global state $\textbf{s}(t)$ or the local observation $\textbf{o}_i(t)$ as the input. Therefore, we generally denote the input as $\textbf{h}_i(t)$ for simplicity of representation. Correspondingly, the division policy $\pi_f$ generates the separation confidence at time step $t$ with ${\textbf{z}}_i(t)=\pi_f\left(\textbf{h}_i(t)\right)$.
Furthermore, agents share the separation confidence with neighbors and concatenate these received values as the enhanced confidence (i.e., ${\textbf{Z}}_i^+(t)$, $i \in\{1, \cdots, N\}$).
After identifying the formation $c$ based on ${\textbf{Z}}_i^+(t)$, agent $i$ reshapes its local observation as $\textbf{o}_{i}^{\prime}(t)=g\left(\textbf{o}_{i}(t),{\textbf{Z}}_i^+(t)\right)$ and computes the action $\textbf{a}_{i}(t)=\pi_{i}^c\left(\textbf{o}_{i}^{\prime}(t)\right)$, so as to realize a new formation $c$ together with new neighbors. Then reward $R(t)$ is computed to evaluate the impact of the joint action $\textbf{a}(t)=\left\{\textbf{a}_{1}(t), \textbf{a}_{2}(t), \cdots, \textbf{a}_{N}(t)\right\}$.
As for distributed system, we let $\epsilon$ to indicate average compound error over time $T$ relative to centralized system as \cite{20122515132876} discussed, while $\epsilon=0$ in centralized scheme.
Here we propose the communication overheads $D(\pi_f, \bm{\pi}_i)$ as the objective of policy optimization, which can be formulated as
\begin{align}
    \label{formulation}
    &\min\limits_{\pi_f,\bm{\pi}_i} D(\pi_f, \bm{\pi}_i) \nonumber\\
    \text { s.t. } &\sum\nolimits_{t=1}^{T} (R(t) - \epsilon) \geqslant r_{\text {thre}}, \nonumber \\
    &\textbf{a}_{i}(t)=\pi_{i}^c\left(g\left(\textbf{o}_{i}(t),{\textbf{Z}}_i^+(t)\right)\right) \\
    &{\textbf{Z}}_i^+(t)=[\textbf{z}_i(t) \|\{\textbf{z}_j(t)|\forall j \in \bm{\xi}_i(t)\}]\nonumber \\     
    &{\textbf{z}}_i(t)=\pi_f\left(\textbf{h}_i(t)\right), \quad    \quad \quad   \quad  \quad \quad \quad \quad \forall i \in\{1, \cdots, N\}\nonumber
\end{align}
where $r_{\text {thre}}$ denotes the required minimum cumulative reward and $\textbf{h}_i(t)= \textbf{s}(t)$, if it adopts a central policy $\pi_f^* $; otherwise, 
$\textbf{h}_i(t)=\textbf{o}_i(t)$, where $\|$ indicates the concatenate operation. 
Notably the communication overheads $D$ heavily depend on the means for controlling (i.e.,  centralized, leader-follower\cite{Najjar2019ALC} and decentralized policies denoted as $\pi_f^*$, $\pi_f^{*'}$ and $\pi_f$). Beforehand, we denote overheads for positions of agent and adversary respectively as $\Omega_{p_i}$ and $\Omega_{p_e}$ while $\Omega_{z}$ indicates the instruction costs and $\Omega_{s}$ for the expenses of sequence number in decentralized system.
\begin{itemize}
    \item \textbf{Centralized}. $N$ positions in $\Omega_{p_i}$-bit of all agents along with $\Omega_{p_e}$-bit adversary's position are needed for centralized controller, based on which it generates instructions in $\Omega_{z}$-bit for each agent. Thus overheads are summarized as $D(\pi_f^{*},\bm{\pi}_i)=N\Omega_{p_i}+\Omega_{p_e}+N\Omega_{z}$ .
    \item \textbf{Leader-Follower}. One agent is selected as the leader to generates $\Omega_{z}$-bit instructions for $N-1$ followers based on $N-1$ positions of followers in $\Omega_{p_i}$-bit while adversary's position is locally obtained. The communication overheads are $D(\pi_f^{*'},\bm{\pi}_i)=(N-1)\Omega_{p_i}+(N-1)\Omega_{z}$. 
    \item \textbf{Decentralized}. Division confidence $\hat{\mathbf{z}}_i(t)$ is computed individually without up-link cost. Along with the $\Omega_s$-bit sequence number, the local confidence $\hat{\mathbf{z}}_i(t)$ in $\Omega_z$ bits is broadcast to $|\bm{\xi}_i(t)|$ neighbors for agent $i$, so the total overheads of fully-decentralized system are $D(\pi_f,\bm{\pi}_i)=\sum_{i\in\mathcal{N}}|\bm{\xi}_i(t)|(\Omega_{s}+\Omega_{z})$.
\end{itemize}
Since the overheads of $\Omega_z$ and $\Omega_s$ are both in bit level while $\Omega_{p_i}$ and $\Omega_{p_e}$ cost few bytes to store positions, the overall overheads can be significantly reduced in distributed execution relative to the other two systems. 
Therefore, we try to develop a decentralized communication-efficient solution.

\section{The Framework of IA-MAPPO}\label{s3}
In this section, we present details of the hierarchical structure IA-MAPPO, in which the low-level mixed-formation policy determines where to go and how to form different formations, and
the up-level division control policy decides when to deconstruct and reform new shapes. Beforehand, we delve into the details of MAPPO \cite{yu2022surprising}, which shall constitute the basis of IA-MAPPO.
\subsection{MAPPO-based Specific Formation Control} \label{single formation policy}
The conventional MAPPO algorithm \cite{yu2022surprising} aims to obtain a  policy $\pi_i^c$ for fixed formation $\mathcal{F}_c$ in FMA. Specifically, MAPPO adopts the CTDE architecture and utilizes importance sampling to stabilize the learning progress. Notably, we assume that all agents are homogeneous in the environment, and thus each agent executes the same individual policy $\pi_{i}^c$ with parameter $\theta$. In particular, MAPPO uses an old-version $\pi_{i, \theta_{\old}}^c$ and value function $V_{\psi_{\old}}$ to interact with the environment, store decision trajectories, and calculate the ratio $\mu_{i}(t)=\frac{\pi_{i,\theta}^c\left(\textbf{a}_{i}(t) \mid \textbf{o}_{i}(t)\right)}{\pi_{i,\theta_{\text {old }}}^c\left(\textbf{a}_{i}(t) \mid \textbf{o}_{i}(t)\right)}$ by the target policy $\pi_{i,\theta}^c$.
Afterwards, parameters $\theta$ and $\psi$ of actor networks and critic networks respectively are periodically updated to maximize
\begin{align}
    J_{\pi_{i}}(\theta)=& \min \left(\mu_{i}(t) \hat{A}(t), \operatorname{clip}\left(\mu_{i}(t), 1-\varepsilon, 1+\varepsilon\right) \hat{A}{(t)}\right), \nonumber \\
    J_{V}(\psi)=& -\left(V_{\psi}\left(\textbf{s}{(t)}\right)-\left(\hat{A}{(t)}+V_{\psi_{\old}}\left(\textbf{s}{(t)}\right)\right)\right)^{2}, \label{MAPPO update}
\end{align}
where $\hat{A}(t)=\sum_{l=0}^{T-t-1}(\gamma \lambda)^{l} \delta{(t+l)}$ is the advantage estimation function at time step $t$ with $\delta{(t)}=R{(t)}+\gamma V_{\psi_{\old}}\left(\textbf{s}{(t+1)}\right)- V_{\psi_{\old}}(\textbf{s}(t))$, within which $R{(t)}$ is defined in \eqref{reward sum}, while $\textbf{s}(t)$ denotes the global state and $\varepsilon$ is a hyperparameter.
Based on the classical MAPPO, the swarm has the capacity to accomplish the function of monitoring the target area and avoiding an adversary in a fixed formation  $\mathcal{F}_c$. 
\subsection{Mixed-Formation Policy}
Following Section \ref{single formation policy}, we can obtain a set of policies respectively for formations $\{\mathbf{\mathcal{F}}_{c}| \forall c \in \mathcal{C}\}$.
In this part, we regard these policies as teacher models (i.e., a teacher model $\pi_{i}^c$ instructs the formation $\mathcal{F}_c$), and utilize policy distillation \cite{green2019distillation} to obtain a mixed-formation policy, so as to reduce local memory occupation.
As shown in Fig. \ref{AI-MAPPO}, we collect both inputs and outputs  of policies $\{\pi_i^c|\forall c \in \mathcal{C} \}$ for multi formation patterns by constituting a replay memory $\mathcal{B}$ as 
\begin{equation}
    \mathcal{B}= \left\langle(\mathbf{o}_{{1}}, \mathbf{a}_{{1}}), \ldots, (\mathbf{o}_{{C}}, \mathbf{a}_{{C}})\right\rangle_{\times \Lambda},
\end{equation}
where $\mathbf{a}_{c}$ is generated by a learned teacher model $\pi_i^{c}$ to form $\mathcal{F}_c$ and $\Lambda$ is the capacity of replay buffer. 
Considering the dimension of observation $\mathbf{o}_{c}$ is determined by agent number $c$, we align the vectors of observations in different formations by zero-padding operation.
During training, memories $\left\langle\mathbf{o}_{{c}}, \mathbf{a}_{{c}}\right\rangle$ from teacher models in buffer are utilized by mixed-formation policy $\pi_{s}$ via minimizing Mean-Squared-Error (MSE) loss
\begin{equation}
\label{distillation update}
\mathcal{L}_{\mathrm{PD}}\left(\theta_{s}\right)=\sum\nolimits_{i=1}^{\Lambda} \sum\nolimits_{c=1}^{C}\left\|\pi_i^{c}(\mathbf{a}_{{c}}|\textbf{o}_{c})-\pi_{s}(\mathbf{a}_{s}|\textbf{o}_{c};\theta_s)\right\|,
\end{equation}
where $\theta_{s}$ is the parameter of $\pi_{s}$, and $\mathbf{a}_{{c}}$ along with $\mathbf{a}_{s}$ refer to the actions specific to the same state $\mathbf{o}_{c}$ respectively from teacher models $\pi_i^{c}$ and the student model $\pi_s$. 
After updating $\theta_{s}$ through \eqref{distillation update}, the mixed-formation policy set $\bm{\pi}_i=\{\pi_i^c | \forall c \in \mathcal{C}\}$ is simplified to one student policy $\bm{\pi}_i=\pi_s$, which significantly saves the usage of agents' memory.

\subsection{Division Control}
With the help of policy distillation, the low-level policy $\pi_s$ is capable to realize $C$ formation patterns, assuming the availability of the teammate matrix $\mathbf{M}_i^{\mathcal{C}}$ of agent $i$, with the size $C\times N$. In other words, agent $i$ can recognize its $c-1$ neighbors when it selects $\mathcal{F}_i(t)=\mathcal{F}_c$ by referring to the teammate vector $\mathbf{m}_i^{c}$ with size $1\times N$ in matrix $\mathbf{M}_i^{\mathcal{C}}$. Therefore, we only need to learn a division control policy $\pi_f$ for formation pattern switching. In particular, the division control consists of policy imitation, confidence communication and masking operation. 

\subsubsection{Policy Imitation}
For a completely centralized system, the controller can adopt a policy $\pi_f^*$ to notify a specific formation pattern $\mathcal{F}_c$ to agent $i$ (i.e., an instruction of a $C$-length one-hot vector $\mathbf{z}_i(t)$ is given that the $c\text{-th}$ element $z_i^c(t)$ is equal to $1$ while others being $0$). In advance, specific intervals $ \delta_{\text {safe}}^c$ are designed for each pattern $c\in \mathcal{C}$. Mathematically,
\begin{equation}
    \label{central control}
    z_i^c(t)= \begin{cases}
    1,& \delta_{\text {min}}^c \le \beta(t) < \delta_{\text {max}}^c; \\
    0, &\text{otherwise},
    \end{cases} 
\end{equation}
where $\beta(t)=\|\frac{1}{|{\mathcal{F}_i}(t)|} \sum_{j \in \mathcal{F}_i(t)}(\mathbf{p}_{j}(t)-\mathbf{p}_{e}(t))\|$ represents the distance between the center of formation $\mathcal{F}_i(t)$ and the adversary $\textbf{p}_e(t) $, while $\delta_{\text {safe}}^c=[\delta_{\text {min}}^c,\delta_{\text {max}}^c)$ denotes the safety distance interval for formation pattern $c$.
Apparently, the centralized control process is energy-consuming in obtaining positions of all agents.

Targeted at getting rid of the central controller and reducing communication overheads, we assume that the featureless agent $i$ obtains $\textbf{h}_i(t)=\textbf{o}_i(t)$ with raw information from neighbors $ \textbf{o}_{\bm{\xi}_i}^{-}(t)$ by communications. After that, the $C$-dimension formation confidence vector $\mathbf{\hat{z}}_i(t)$ is generated by a decentralized policy network $\pi_{f}$ parameterized by $\theta_f$ which is shared among agents, denoted as
\begin{equation}
\mathbf{\hat{z}}_i(t)=\pi_{f}(\textbf{h}_i(t)),
\end{equation}
the element of which $\hat{z}_i^c(t)$ represents the derived confidence to pattern $c$ for agent $i$ in a distributed manner.

We adopt IL to update the parameter $\theta_f$. Specifically, in the data-collecting phase, $\mathbf{z}_i(t)$ is computed by the central policy according to \eqref{central control} and then the state-action pairs $(\textbf{h}_i(t),\mathbf{z}_i(t))$ are stored at each time-step $t$.
In the training phase, we define $\mathbf{z}_i(t)$ as the label of $\mathbf{\hat{z}}_i(t)$ and compute the MSE loss as
\begin{equation}
    \mathcal{L}(\theta_f)= \sum\nolimits_{t=0}^{T} \sum\nolimits_{i\in \mathcal{N}}\|\pi_{f}\left (\textbf{h}_{i}(t);\theta_f \right) -\mathbf{z}_i(t)\|,
\end{equation}
where $T$ is the length of each episode.
Thus an estimator of global controller $\mathbf{\hat{z}}_i(t)$ can be attained in a decentralized manner from $\pi_f$. 

\subsubsection{Confidence Communication}
To enhance the cooperation and stability of the system, it is essential for agents to communicate their separation confidence to understand each other. After calculating $\mathbf{\hat{z}}_i(t)$ by $\pi_f$, agent $i$ broadcasts the result to neighbors in $\delta_\text{com}$. Meanwhile, it receives neighbors' confidence collected as a vector $\{\mathbf{\hat{z}}_j(t)|\forall j \in \bm{\xi}_i(t)\}$, denoted as $\hat{\mathbf{Z}}_{-i}(t)$.
Up to this point, the total division confidence that agent $i$ can partially gather is $\mathbf{\hat{Z}}_i^+(t)=[\mathbf{\hat{z}}_i(t) \|\hat{\mathbf{Z}}_{-i}(t)]$. 

\subsubsection{Masking Operation}
The masking operation transfers the raw observation $\textbf{h}_i(t)$ into $\textbf{h}^{\prime}_i(t)$ to meet the input of low-level mix-formation policy $\pi_s$,
 and it involves three sub-procedures.
\begin{itemize}
    \item \textbf{Information Aggregation}.  $\mathbf{\hat{Z}}_i^+(t)$ contains all neighbors' intentions of which formation pattern to choose at next time step. Recalling the definition of $\hat{\mathbf{Z}}_{-i}(t)$ and $\mathbf{\hat{z}}_i(t)$, $\mathbf{\hat{Z}}_i^+(t)$ is a matrix with the size of $|\bm{\xi}_i(t)+1| \times C$. In order to aggregate preferences in the domain $\bm{\xi}_i(t)$ of agent $i$, we apply column summation to $\mathbf{\hat{Z}}_i^+(t)$ to figure out the importance of each formation pattern
    \begin{equation}
        \phi_i^c(t) = \sum\nolimits_{j \in \{i \cup \bm{\xi}_i(t)\}}\hat{z}_j^c.
    \end{equation}
    Correspondingly, we further obtain a $C$-length importance vector $\bm{\phi}_i(t)$.
    \item \textbf{Formation Pattern Determination}. After obtaining the importance vector $\bm{\phi}_i(t)$, we determine the formation pattern $c^\sharp$ for next time-step $t+1$ by locating the index $c^\sharp$ corresponding to the maximum value in $\bm{\phi}_i(t)$.
    
    \item \textbf{Observation Reshape}. As preliminary assumed, agent $i$ refers the set of teammates $\mathbf{m}_i^{c^\sharp}$ from the teammates matrix $\mathbf{M}_i^{\mathcal{C}}$ in its memory for current pattern $c^\sharp$.
    Then it computes $\textbf{h}_i(t)$ and updates
    \begin{equation}
        \textbf{h}^{\prime}_i(t)=[\textbf{o}_i^{-}(t) \| \{\textbf{o}_j^{-}(t) | \forall j \in \mathbf{m}_i^{c^\sharp}\}].
    \end{equation}
    Notably, due to the adoption of IL from central policy, the precondition $\mathbf{m}_i^{c^\sharp} \subseteq \bm{\xi}_i(t)$ of transition $\textbf{h}^{\prime}_i(t) \gets \textbf{h}_i(t)$ is always satisfied.
\end{itemize}

\subsection{Alternative Training} \label{alternative training}
The IL-induced decentralized division policy $\pi_{f}$ degrades the performance relative to central policy $\pi_f^*$ because of compounding deviations, which leads to the deterioration of $R(t)$ as \eqref{formulation}. Theoretically, it is proved in \cite{20122515132876} as
\begin{equation}
    \label{compounding error}
    \mathbb{E}_{s \sim d_{\pi_{f}}}\left(C_{\pi}(s)\right) \le  \mathbb{E}_{s \sim d_{\pi_f^*}}\left(C_{\pi}(s)\right)+T^{2} \epsilon,
\end{equation}
where $C_{\pi}(s)$ denotes the expected episodic cost of performing policy $\pi$ in state $s$. The distribution $d_{\pi}$ encodes the state visitation frequency over $T$ steps with policy $\pi$, and $\epsilon$ represents the average compounding deviations over time $T$ between $\pi_{f}$ and $\pi_f^{*}$. As indicated in \eqref{compounding error}, the extra cost for imitative policy $\pi_{f}$ grows quadratically and can result in tremendous performance loss.

To rectify the adverse effects incurred by IL, we leverage AT by fine-tuning mixed-formation policy $\pi_s$ after learning the division policy $\pi_f$. Therefore, through interacting with the environment by the joint actions $\mathbf{u}_{t}$ of division instructions $\hat{\textbf{z}}_i(t)$ and acceleration actions $\textbf{a}_i(t)$ respectively from $\pi_{f}$ and $\pi_{s}$, we first collect trajectories 
$\left\langle(\mathbf{o}_{1}, \mathbf{u}_{1}), \ldots, (\mathbf{o}_{T}, \mathbf{u}_{T})\right\rangle$. Different from manual trajectory labeling in \cite{20122515132876}, we design a task-oriented alternative training reward $R_{\at}$ to evaluate the state $\mathbf{o}_{t}$.
Specifically, in order to rectify the undesired situations (e.g., the reluctancy to unite due to the over-long inter-group distance), $R_{\at}$ is designed as 
\begin{equation}
    \label{vf}
    R_{\at}(\mathbf{o}_{t}) = - \sum\nolimits_{i \ne j}\|\bar{\mathbf{p}}_{q_i}(t) -\bar{\mathbf{p}}_{q_j}(t)\|,
\end{equation}
where $q_i$ and $q_j$ are two arbitrary groups in $\{q_1,\dots,q_{\chi(t)}\}$, and both $\bar{\mathbf{p}}_{q_i}$ and $\bar{\mathbf{p}}_{q_j}$ signify the center of them.
\eqref{vf} evaluates $\mathbf{o}_{t}$ by computing average distance between two groups. Afterwards, we transfer the tuples into trajectories $\left\langle(\mathbf{o}_{1}, \mathbf{u}_{1},\mathbf{o}_{2},R_{\at}(\mathbf{o}_{1})), \ldots, (\mathbf{o}_{T}, \mathbf{u}_{T},\mathbf{o}_{T+1},R_{\at}(\mathbf{o}_{T}))\right\rangle$ and fine-tune the parameter $\theta$ of $\pi_s$ according to \eqref{MAPPO update}.

\section{Experimental Results and Discussions}\label{s4}
In this section, we evaluate the performance of IA-MAPPO in multi-agent particle environment \cite{lowe2017multi} for FMA task, so as to verify the reduction in the communication overheads, on the basis of competitive performance as the centralized system.

Beforehand, we select RAN-MAPPO, CE-MAPPO, AT-MAPPO, LF-MAPPO, and IL-MAPPO as baselines. In particular, RAN-MAPPO refers to the algorithm with randomly generated division instructions on top of a mixed-formation policy $\pi_s$, while CE-MAPPO indicates the centralized division policy $\pi_f^*$ with  $\pi_s$. On top of CE-MAPPO, AT-MAPPO denotes the CE-MAPPO with $\pi_s$ being further fine-tuned via AT, while  
LF-MAPPO substitutes the leader-follower policy $\pi_f^{*'}$ for $\pi_f^{*}$ in CE-MAPPO. IL-MAPPO implies the absence of AT in IA-MAPPO. Components of these algorithms are summarized in Table \ref{comparison}.

In the simulation, we focus on the FMA task that requires transformation between $\mathcal{F}_8$ in low velocity $v_{\text{unite}}$ and $\mathcal{F}_4$ in high velocity $v_{\text{split}}$ with limited communication range $\delta_\text{com}$ for each agent. The swarm is expected to avoid attacks from the adversary and stay in a target area with radius $r_{\mathcal{T}}$. An adversary keeps pursing at the speed of $v_{\text{adversary}}$, whose policy is chasing the center of swarm all the time. 
The performance of system is quantified by the reward defined in \eqref{total rewards}. The key parameters are summarized in Table \ref{tab:para}.
\begin{table}[!b]
    \vspace{-2.5em}
    \centering
    \caption{Component comparison of algorithms.}
    \label{comparison}
    \vspace{-1em}
    \begin{tabular}{|c|c|c|c|c|c|c|}
    \hline
        Name(-MAPPO) & RAN & CE & LF & AT & IL & IA \\ \hline
        Decentralized & \Circle & \Circle & \RIGHTcircle & \Circle & \CIRCLE & \CIRCLE \\ \hline
        With AT& \Circle & \Circle & \Circle & \CIRCLE & \Circle & \CIRCLE  \\ \hline
    \end{tabular}
\end{table}
\begin{table}[]
    \vspace{-2em}
    \centering
    \caption{The remaining key parameter settings.}
    \label{tab:para}
    \begin{tabular}{|l|l|}
    \hline
     \multicolumn{1}{|c|}{\textbf{Parameters}} & \multicolumn{1}{c|}{\textbf{Value}} \\ \hline 
     \multicolumn{1}{|c|}{Number of Agents and Adversary}      & \multicolumn{1}{c|}{$n_a=8, n_e=1$}    \\ \hline
      \multicolumn{1}{|c|}{Discount Factor}    & \multicolumn{1}{c|}{$\gamma=0.8$} \\ \hline
     \multicolumn{1}{|c|}{Communication Range}      & \multicolumn{1}{c|}{$\delta_\text{com}=2\  \text{m}$}    \\ \hline
      \multicolumn{1}{|c|}{Radius of Target Area}      & \multicolumn{1}{c|}{$r_{\mathcal{T}}=8\   \text{m}$}    \\ \hline
    \multicolumn{1}{|c|}{Time Steps each Episode}      & \multicolumn{1}{c|}{$T=200\ \text{s}$}     \\ \hline
   \multicolumn{1}{|c|}{Agent Number of each Formation}      & \multicolumn{1}{c|}{$c_1=8,c_2=4$}    \\ \hline
      \multicolumn{1}{|c|}{Safe Distance}   
     & \multicolumn{1}{c|}{$\delta_{\text {safe}}^{4}=[0,2.8),\delta_{\text {safe}}^{8}=[2.8,\infty )\ \text{m}$}    \\ \hline
    \multicolumn{1}{|c|}{Parameters $(k_f,k_a,k_e,\varphi_f, \alpha, \varphi_m)$}      & \multicolumn{1}{c|}{$(1,0.3,20,0.75,0.5,100)$}  \\  \hline
    \multirow{2}{*}{\makecell[c]{\ \ \ \ \ \  Maximum of Velocity}} 
     &\makecell[c]{$v_{\text{unite}}=0.5\ \text{m/s}, v_{\text{split}}=1.2\ \text{m/s},$}\\
     &\makecell[c]{$v_{\text{adversary}}=0.6\ \text{m/s}$} \\ \hline
    
    \end{tabular}

\end{table}

\subsection{Communication Overheads} \label{5}
Table \ref{tab:ptphd} concludes the communication costs of six systems that respectively represents centralized, leader-follower and decentralized architectures, when $[\Omega_{p_i},\Omega_{p_e}, \Omega_{z}, \Omega_{s}]$ is set as $[64, 64, 1, 3]$ bits, since a single-bit $\Omega_{z}$ is required due to $C=2$ and $3$-bit $\Omega_{s}$ stands for sequence number as $n_a=8$. Besides, $\Omega_{p_i}$ and $\Omega_{p_e}$ represent $(p_x,p_y)$ of agents and adversary in $32$-bit floating-point format.
It is consistent with the analysis in Section \ref{problem formulation} that the decentralized IL/IA-MAPPO decrease the communication overheads respectively to $34.1\%$/$39.2\%$ of CE-MAPPO and $44.0\%$/$50.6\%$ of LF-MAPPO.


\begin{table}[]
\centering
\vspace{-1.5em}
\caption{Average communication overheads each episode in typical systems for division control.}
\label{tab:ptphd}

\begin{tabular}{|l|l|l|l|}
\hline

\makecell[c]{\textbf{Methods}} & \makecell[c]{\textbf{Up-link}} & \makecell[c]{\textbf{Down-link}} & \makecell[c]{\textbf{Overall Cost}}  \\ 
\hline

\multirow{1}{*}{\makecell[c]{CE/RAN-MAPPO}}
& \makecell[c]{$11.7\ \text{KB}$}    
& \makecell[c]{$1.44\ \text{KB}$}    
& \makecell[c]{$13.14\ \text{KB}$}       \\
\hline

\multirow{1}{*}{\makecell[c]{\ \ \ \ AT-MAPPO}}
& \makecell[c]{$12.47\ \text{KB}$}  
& \makecell[c]{$1.53\ \text{KB}$ }  
& \makecell[c]{$14.0\ \text{KB}$}    \\

\hline
\multirow{1}{*}{\makecell[c]{\ \ \ \ LF-MAPPO}}
& \makecell[c]{$10.02\ \text{KB}$ }  
& \makecell[c]{$0.15\ \text{KB}$ }  
& \makecell[c]{$10.18\ \text{KB}$}     \\
\hline
\multirow{1}{*}{\makecell[c]{\ \ \ \ IL-MAPPO}}
& \makecell[c]{$-$}  
& \makecell[c]{$4.48\ \text{KB}$ }  
& \makecell[c]{$4.48\ \text{KB}$}    \\
\hline
\multirow{1}{*}{\makecell[c]{\ \ \ \ IA-MAPPO}}
& \makecell[c]{$-$}  
& \makecell[c]{$5.15\ \text{KB}$ }  
& \makecell[c]{$5.15\ \text{KB}$}    \\
\hline
\end{tabular}
\vspace{-1.5em}
\end{table}

\subsection{Pursuit Avoidance Performance}

\begin{figure}[t]
    \centering
    \vspace{-1em}

    \includegraphics[width = 0.32\textwidth]{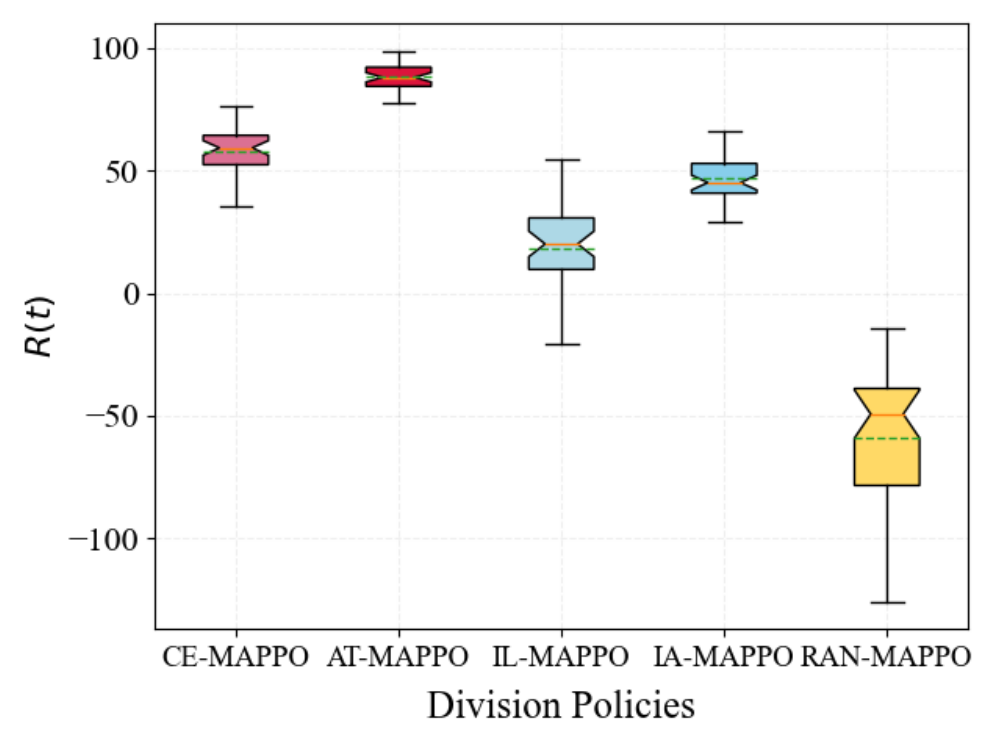}
    \vspace{-0.8em}

    \caption{Average $R(t)$ over $600$ episodes in five division policies.}
    \vspace{-0.5em}
    \label{3}
\end{figure}

\begin{figure}[tbp]
    \vspace{-0.2em}
    \label{4}
    \centering
    \subfigure[\text{Event counts}]{
    \includegraphics[width = 0.231\textwidth]{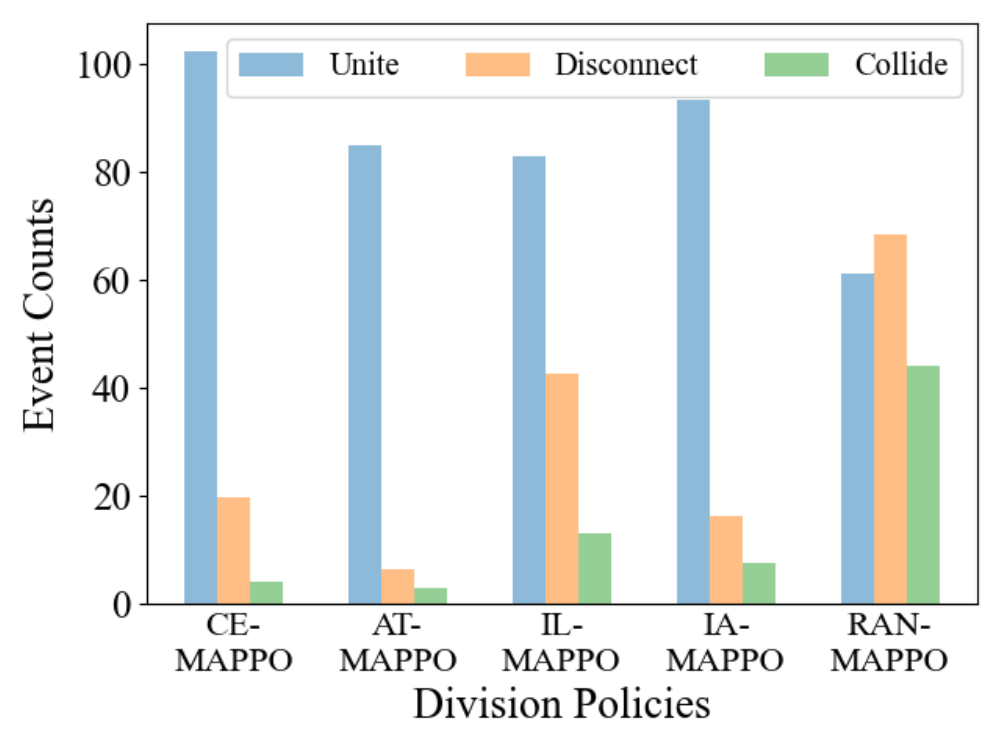} \label{cl3}}
    \subfigure[\text{Variation of $R_{\at}(t)$}]{
    \includegraphics[width = 0.231\textwidth]{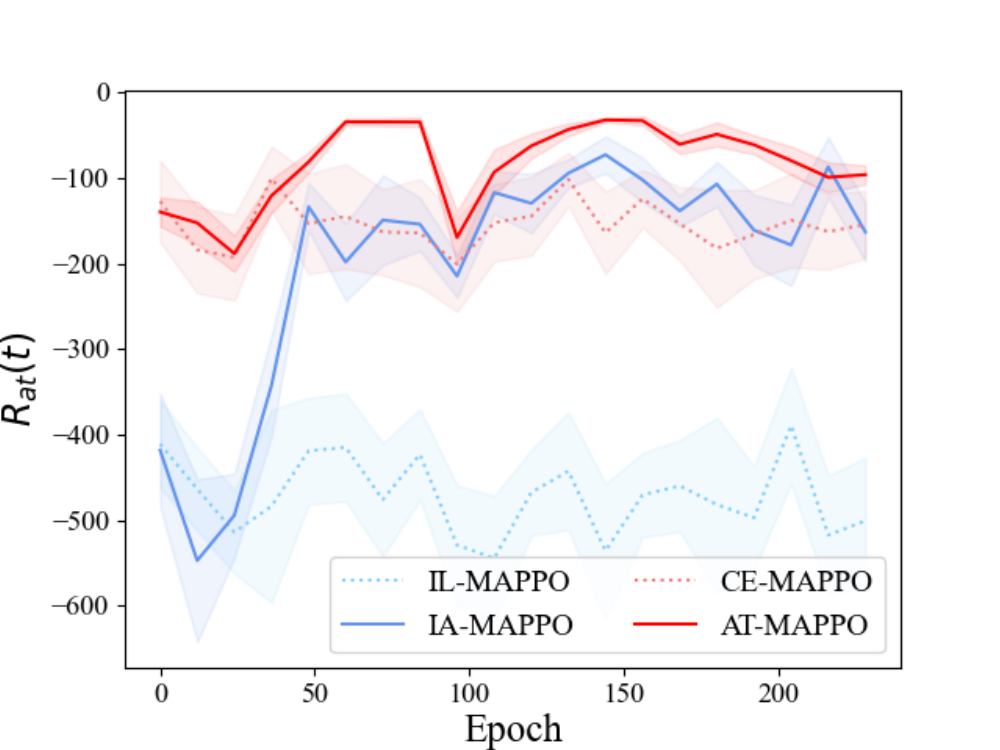}\label{cl2}}
    
    \caption{Average event counts over $600$ episodes and variation of $R_{\at}(t)$ during fine-tuning CE-MAPPO and IL-MAPPO.}
    \vspace{-1.5em}
\end{figure}

Fig. \ref{3} presents the pursuit avoidance performance of IA-MAPPO and other algorithms. 
Notably, RAN-MAPPO exhibits worst performance due to severe communication disconnections and collisions with the adversary in Fig. \ref{cl3}.
It can be observed that CE-MAPPO yields evidently better than the IL-MAPPO in Fig. \ref{3}, while it costs heavier communication overheads as Table \ref{tab:ptphd} presents. Consistent with the discussion in Section \ref{alternative training}, the distributed system (IL-MAPPO) significantly reduces the communication expenses at the sacrifice of connection and safety maintenance in Fig. \ref{cl3}.

With the implementation of AT, IA-MAPPO obtains comparable performance as CE-MAPPO in Fig. \ref{3}, as collisions with adversary occur less frequently in IA-MAPPO with plunging disconnections between agents and encouraged intention sharing, and similar observations also applies to the centralized AT-MAPPO shows in Fig. \ref{cl3}. 
Specifically, the superiority of AT can be explained as the reflection of the improved $R_{\at}$ in the process of fine-tuning the mixed-formation policy $\pi_f$ in Fig. \ref{cl2}. 

\section{Conclusion}\label{s5}
In this paper, we have proposed IA-MAPPO to solve the problem of FMA. Specifically, we have enhanced the flexibility of formation by distilled policies and utilize Imitation Learning to obtain a decentralized solution with significantly reduced communication overheads. Afterwards, Alternative Training is put forward to complement the performance loss incurred due to the decentralization. Finally, we have verified the communication efficiency and proven the effectiveness of IA-MAPPO in extensive experiments.
In the future, we will enlarge the scale of swarms and enrich formation patterns. More practical factors (e.g., communication delay) shall be accurately incorporated as well.

\bibliographystyle{IEEEtran}
\vspace{-0.2em}

\bibliography{reference}
\end{document}